\documentclass{article}

\usepackage[preprint]{neurips_2022}

\usepackage[utf8]{inputenc} 
\usepackage[T1]{fontenc}    
\usepackage{hyperref}       
\usepackage{url}            
\usepackage{booktabs}       
\usepackage{amsmath,amssymb,amsfonts}       
\usepackage{nicefrac}       
\usepackage{microtype}      
\usepackage{xcolor}         
\usepackage{graphicx}
\usepackage{subfig}

\title{HyBNN and FedHyBNN: (Federated) Hybrid Binary Neural Networks}

\author{%
  Kinshuk Dua\\
  School of Computer Science and Engineering\\
  Vellore Institute of Technology\\
  Chennai, India\\
  \texttt{kinshukdua@gmail.com}\\
}

\begin{document}

\maketitle

\begin{abstract}
Binary Neural Networks (BNNs), neural networks with weights and activations constrained to -1(0) and +1, are an alternative to deep neural networks which offer faster training, lower memory consumption and lightweight models, ideal for use in resource constrained devices while being able to utilize the architecture of their deep neural network counterpart. However, the input binarization step used in BNNs causes a severe accuracy loss. In this paper, we introduce a novel hybrid neural network architecture, Hybrid Binary Neural Network (HyBNN), consisting of a task-independent, general, full-precision variational autoencoder with a binary latent space and a task specific binary neural network that is able to greatly limit the accuracy loss due to input binarization by using the full precision variational autoencoder as a feature extractor. We use it to combine the state-of-the-art accuracy of deep neural networks with the much faster training time, quicker test-time inference and power efficiency of binary neural networks. We show that our proposed system is able to very significantly outperform a vanilla binary neural network with input binarization. We also introduce FedHyBNN, a highly communication efficient federated counterpart to HyBNN and demonstrate that it is able to reach the same accuracy as its non-federated equivalent. 
We make our source code, experimental parameters and models available at: https://anonymous.4open.science/r/HyBNN.
\end{abstract}

\section{Introduction}

Deep Neural Networks (DNNs) have been extensively applied to an incredibly wide range of applications due to their ability to generate remarkable results on complex tasks once thought to be almost impossible such as face recognition, self-driving systems, natural language processing (NLP), image analysis, weather prediction etc \cite{liu2017survey}. They form the basis for most state-of-the-art machine learning algorithms and are even capable of outperforming humans in some areas \cite{yu2017sketch}.

Deep Neural Networks draw their success from being able to have hundred to thousands of layers consisting of millions or even billions of trainable parameters. For instance, GPT-3 has over 175 billion parameters that have been estimated to cost over \$12 million to train \cite{floridi2020gpt}. This is because training a DNN i.e. Deep Learning requires a long chain of expensive matrix and tensor computation consisting of millions of floating-point operations. 

Despite their success, a lot of resource constrained devices, such as mobile phones, IoT devices, embedded systems and other edge devices do not have the necessary computing capacity in terms of available processing power, memory, bandwidth or energy to sufficiently and quickly train a DNN. As such, training these networks require specialized high-performance hardware, like GPUs, which are expensive and consume an enormous amount of energy to operate. 

Optimizing deep learning algorithms to be lightweight enough for resource-constrained devices is still an open and active research problem. There have been numerous proposals for the same, including straightforward solutions like training smaller models, or more complex ones like knowledge distillation \cite{hinton2015distilling}, parameter quantization \cite{gong2014compressing}, parameter pruning \cite{han2015learning}, transferred convolution filters \cite{zhang2018shufflenet} and low-rank parameter factorization \cite{wen2018adaptive}. 

Parameter Quantization is especially interesting, since it can greatly reduce the size and computations required for DNNs quickly and lets us reuse existing neural network architectures as is. Binary or Binarized Neural Networks (BNNs) are an extreme form of quantization with weights and activation constrained to -1(0) or +1 at run-time and for gradient computation during training \cite{hubara2016binarized}. 

Since all the weights and activations are binary, the run-time can be heavily reduced as the computationally expensive matrix operations can be replaced with simple XNOR and popcount operations \cite{rastegari2016xnor}. These low-level operations are highly optimized on most modern processors and can even be applied in parallel on processors that support SIMD (Single Instruction, Multiple Data) instructions. Furthermore, these networks can be easily constructed using electronic circuits via XNOR logic gates, greatly improving performance at a fraction of the cost of a GPU. There are however, challenges that limit the widespread adoption of BNNs in real world scenarios. Since the inputs are real-valued they must be converted to binary using a binarization function, the most common is the $sign$ function.

\[ sign(x) = \begin{cases} 
          +1, & \text{if } x\geq 0 \\
          -1, & otherwise 
   \end{cases}
\]
    
The problem with the binarization step is that a lot of information might be lost due to the quantization.

Moreover, the derivative of the sign function is zero which means traditional methods of training neural networks like Stochastic Gradient Descent (SGD) cannot be used to train binary neural networks, fortunately a lot of work has been published about efficient training algorithms for Binary Neural Networks to overcome this problem.

Binary neural networks have been extensively implemented on Field-Programmable Gate Arrays (FPGA), as they can directly compute XNOR operations via logic gates at relatively low computation and energy costs. Even implementation on non-specialized hardware, like CPUs have been proved to decrease computation time memory consumption by a huge margin \cite{rastegari2016xnor}. However, even with all its benefits Binary Neural Networks do come at the cost of an accuracy drop, for example the recent XNOR-Net++ showed a drop of $\approx$ 18\% compared to a real-valued DNN with the same architecture \cite{bulat2019xnor}.

In this paper, we propose and implement a hybrid full-precision binary neural network and its communication-efficient federated counterpart that is able to improve the performance of binary neural networks by using extracting features from the input image using a binary autoencoder trained on a large-scale general dataset. These binary features are then fed directly to the binary neural network, thus eliminating the problem of the input binarization loss. We exploit the fact that a single forward pass for inference in Deep Neural Networks is orders of magnitudes faster than training one from scratch. While training on Binary Neural Networks is extremely quick thanks to their ability to be reduced to simple, highly optimized XNOR and popcount operations and the small size of the parameters. We combine the incredible accuracy of Deep Neural Networks and the extremely lightweight nature and quick training time of Binary Neural Networks to create a hybrid full-precision binary neural network that is able to easily outperform a binary neural network on the same task. 

The contribution of this paper is threefold- 
\begin{enumerate}
    \item First, we present a method to minimize the input binarization loss for the binary neural networks.
    \item Second, we are, to our knowledge, one of the first to propose and implement a hybrid network consisting of both a full-precision and a binary neural network
    \item Lastly, we are, to our knowledge, the first ones to implement a federated binary neural network.
\end{enumerate}

\subsection{Transfer Learning}
In transfer learning, the network is trained on a huge general dataset to learn the representation of the input and then used as a starting point to fine tune on domain specific data. It can be used to transfer knowledge across domain, especially in cases where the domain specific dataset might not be large enough to sufficiently train a deep model with required accuracy.

Formally, given a specific learning task $\mathcal{T}_s$ based on $\mathcal{D}_s$ and another task $\mathcal{T}_g$ based on $\mathcal{D}_g$, Transfer learning is the process of trying to improve the performance of a predictive function $f_\mathcal{T}(\cdot)$ for the task $\mathcal{T}_s$ by
discovering and transferring latent knowledge from $\mathcal{D}_g$ and $\mathcal{T}_g$, where $\mathcal{D}_s \neq \mathcal{D}_g$ and/or $\mathcal{T}_s \neq \mathcal{T}_g$. In addition, in most cases $size(\mathcal{D}_g)\gg size(\mathcal{D}_s)$ .

Specifically network-based transfer learning focuses on transferring some layers of a network trained on the task $\mathcal{T}_g$ with data $\mathcal{D}_g$, including the architecture and parameters to a new deep neural network which can be used for a domain specific task $\mathcal{T}_s$. It is based on the assumption that the frontal part of a network can be used as a feature extractor to learn the general representation of the data which can then be used to improve the performance for the new task $\mathcal{T}_s$ \cite{tan2018survey}. The layers that are transferred over are said to be frozen, meaning that their parameters are cannot be changed during training.

In our case, a variational autoencoder (VAE) with a binary latent space is used as a feature extractor which is trained on a public dataset and then frozen so that the rest of the network can be trained on the given task.

\subsection{Variational Autoencoders}
Variational Autoencoders are neural networks that are used to learn a multivariate latent encoding variable model by maximizing the marginal likelihood of the input data. Since the true posterior is usually intractable, an approximate posterior function is used in practice instead to speed up computations.

They consist of an encoder, which compresses unlabelled data into a constrained latent variable model, known as the latent space representation and a decoder which reconstructs the input data using the latent space. 

By training to reduce reconstruction loss, variational autoencoders are able to reduce the dimensionality of the data and learn efficient coding of the input in the latent space.

Variational autoencoders have been used for a variety of tasks such as generative models alongside Generative Adversarial Networks (GANs), face recognition, representation learning, feature detection etc.

\begin{figure}[h]
\centering
\includegraphics[width=0.46\textwidth]{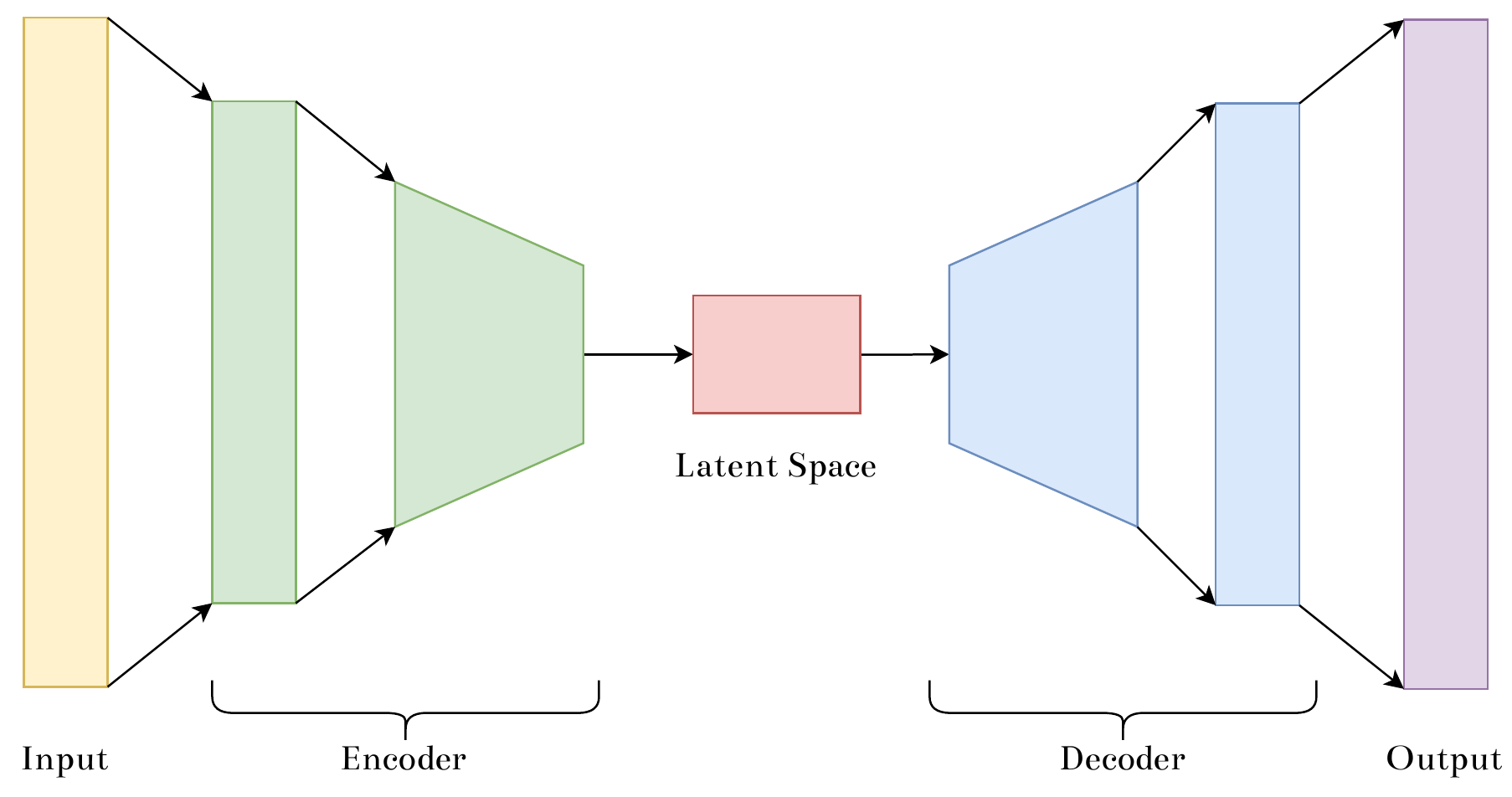}
\caption{Architecture of a Variational Autoencoder}
\end{figure}

\section{Related Work}
\label{sec:related_work}

Binary Neural Networks were first described by Hubara et al. in 2016 \cite{hubara2016binarized}, the paper described the architecture, the binarization process, and a baseline implementation. They solved the problem of the zero-gradient by propagating gradients through discretization and using STE (Straight Through Estimators). 

A lot of work has been done on specifically improving the binarization (quantization) process since it is one of the main source of information loss \cite{qin2020binary}.

Soon after the original binary neural network paper, Rastegori et al. \cite{rastegari2016xnor} showcased XNOR-Net, the binary counterpart to AlexNet, which was used to produce state-of-the-art results with 52x faster execution and 32x less memory consumption in the ImageNet task . It included all methods described in the original binary neural network paper but also proposed an addition "gain" term to counter the loss of data during the binarization step.

Wide Reduced-Precision Network proposed by Mishra et al. \cite{mishra2017wrpn} also showcased a similar way to reduce binarization error but also recommended increasing the amount of filters present in each layer to increase the accuracy of the network. Hu et al. \cite{hu2018hashing} re-framed the binarization process as a hashing problem, leading to a significant increase in accuracy.

Instead of the original one step binarization introduced in XNOR-Net, High-Order Residual Quantization (HORQ) \cite{li2017performance} used a recursive method to produce a series of binary version of the input each with a decreased magnitude scaling factor. ABC-Net by Lin et al. \cite{lin2017towards} took a similar approach and approximated real-valued weights by combining many binary weights and using multiple activation functions. IR-Net designed by Qin et al.\cite{qin2020forward} used a new kind of binarization where the data distribution is reshaped before the binarization process leading to less loss of data. To increase the range of trainable values, Xu et al. \cite{xu2019accurate} proposed using different scaling factors for weights and for activation.

Yang et al. \cite{yang2019quantization} were the first to use a differentiable non-linear quantization function, marking a step away from the $sign$ function used in the original paper. This also let them use more conventional training algorithms like SGD.
Zhou et al. in their DoReFa-Net \cite{zhou2016dorefa} provided a way to increase the accuracy of the weights and activations by using low bitwidth parameter gradients.

Binary autoencoders are autoencoders that map the input to a low-dimensional binary latent space, i.e. the code layer is binary \cite{carreira2015hashing}.
Deng et al. \cite{deng2010binary} were one of the first ones to use deep autoencoders for the purpose of binary coding. Specifically, it was used to encode speech spectrograms. 

The most common application of binary autoencoders however has been for hashing. It has been successfully applied on images \cite{carreira2015hashing} and on videos \cite{song2018self} to produce state-of-the-art results.
Recently, binary autoencoders have been used for continual learning, specifically Deja et al. \cite{deja2020binplay} used them to create unique sequential binary codes for each input sample, similar to hashing and then using older hashes for generative replay to solve the problem of catastrophic forgetting. 

The work on transfer learning with binary neural networks is scarce. Most of the published work talk about improving the performance of binary networks by transfer learning using full-precision networks.
For instance Xie et al. \cite{xie2020deep} devised deep transferring quantization (DTQ), a learnable attentive transfer
module to train binary neural networks using pre-trained full precision models and transfer only relevant knowledge representation. \cite{livochka2021initialization} showcased a similar method to train binary neural networks using full-precision DNNs as a starting point.

Leroux et al. in 2017 \cite{leroux2017transfer} were the first to demonstrate that it was effective to also apply transfer learning in a similar way to DNNs. They also illustrated that it was possible to have a hybrid software-hardware transfer learning setup where the last layer is software based and retrained while the front of the network is constructed in hardware and hence frozen. We however use a different approach, where the first few layers are full-precision (FP) and the rest form the binary neural networks, the reason for doing that will become apparent when we go through the design of our proposed model.

\section{Hybrid BNN (HyBNN)}
\begin{figure}[h]
\centering
\includegraphics[width=0.6\textwidth]{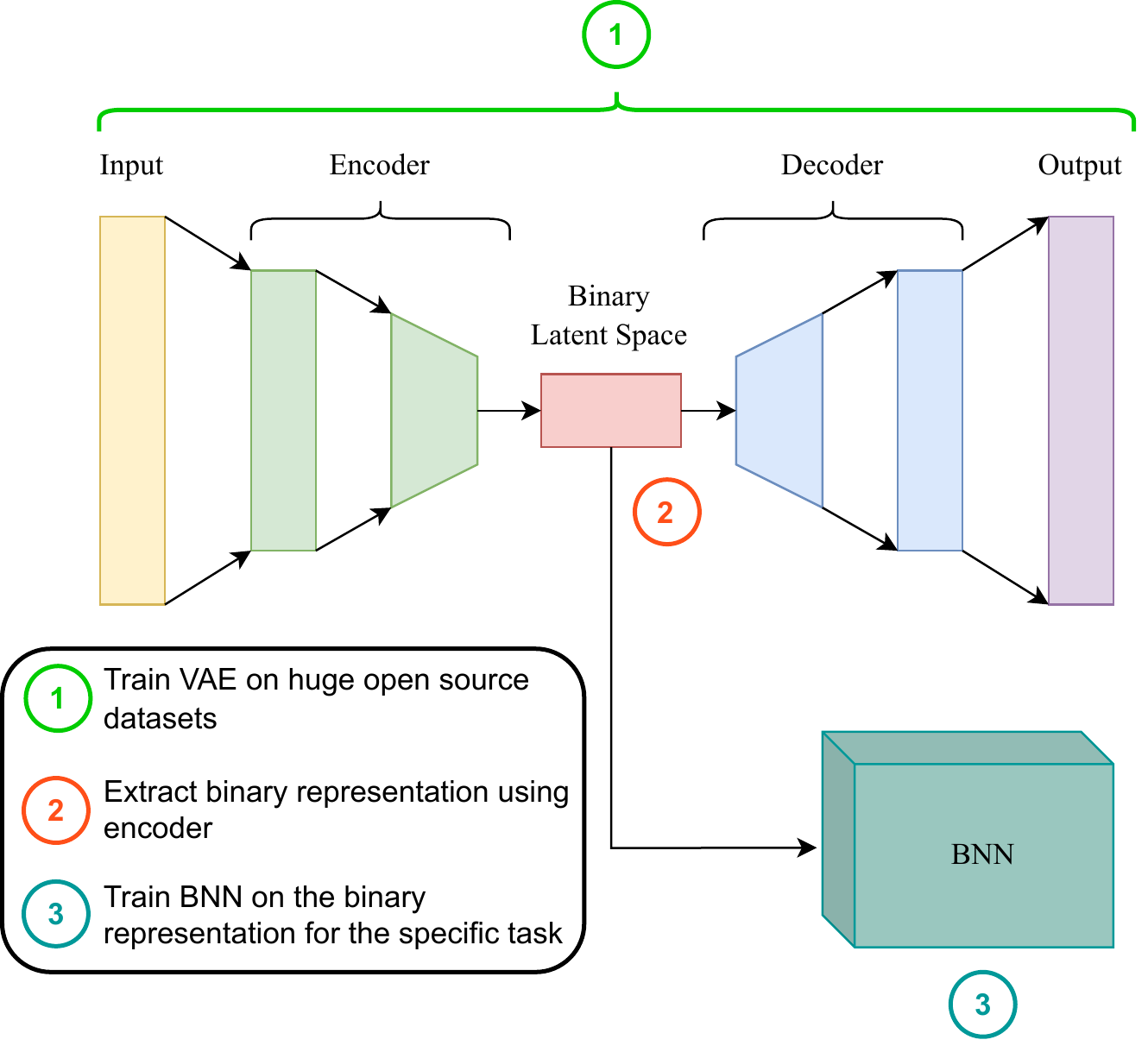}
\caption{Architecture of HyBNN}
\label{fig:hybnn}
\end{figure}

In this section, we will explain the architecture of our proposed system: Hybrid Binary Neural Network (HyBNN) and how it follows naturally from the major takeaways in the literature review of BNNs described in Section \ref{sec:related_work}. The diagram of the architecture is given in Figure \ref{fig:hybnn}.

Takeaways:
\begin{enumerate}
    \item Binary neural networks are extremely lightweight and fast, but suffers from accuracy loss.
    \item Input binarization is a major source of information loss and an open problem.
    \item Binarization can be re-framed as a hashing problem.
    \item Variational autoencoders with binary latent space (binary autoencoders) can be used for hashing.
    \item Hybrid software-hardware transfer learning can increase performance.
\end{enumerate}

The proposed system consists of two key networks- 1) A general full-precision Variational Autoencoder with a binary latent space, and 2) A task specific Binary Neural Network. The full-precision autoencoder works as a binary feature extractor. The extracted features are then used as the input for the Binary Neural Network.
To solve the problem of input binarization loss we use the Variational Autoencoder with a binary latent space as described in BinPlay \cite{deja2020binplay} to extract the binary representation of the input. This input can then be directly fed to the binary neural network, bypassing the initial binarization step entirely. This serves to not only increase the accuracy but also reduces the size of the trainable parameters. The encoder can be trained on a huge general dataset and then frozen. Then the task specific binary neural network can be trained using the output of the encoder as the input. The encoder in the variational autoencoder is used as a general binary feature extractor.
This creates a hybrid full-precision and binary neural network that can be trained on a general dataset and then transfer learned to a specific task, increasing the accuracy to comparable levels to full precision networks while being the fraction of a size.  The size reduction not only helps with faster training and inference but is also ideal for implementation in federated learning as described in Section \ref{sec:FedHyBNN}. The autoencoder can also be reused for multiple different tasks as well since the encoder is only used for feature extraction while the Binary Neural Network can be tweaked for each seperate task. The frozen VAE can be implemented in software while the BNN can be implemented in hardware increasing the efficiency of the system.

The HyBNN training can be broken down into three key steps:

\begin{enumerate}
    \item Representation Learning Phase.
    \item Binary Feature Extraction phase.
    \item Training Phase.
\end{enumerate}

\subsection{Representation Learning Phase}
\label{sec:RLP}
The first phase is the representation learning phase which takes place in a server with good computing power. In this phase, the variational autoencoder is first trained on a large-scale generalized dataset then frozen. This phase is important as the variational autoencoder learns the binary latent representation of the input so that it can be used as a binary feature extractor later on. For example, in case of image classification the autoencoder can be trained on a massive, publicly available dataset such as COCO or CIFAR-100. A large-scale dataset is necessary to make sure that the encoder generalizes well. Since, the encoder is frozen after the pre-training phase, the right selection of the dataset ensures the reliability of the binary representation of the input data during the training step.
After being frozen only the encoder part of the autoencoder is optimized transferred to edge devices, where it is required only for inference.

This is especially important in a federated ecosystem, since the data that is used to train the binary neural network cannot be transferred out of the device for training and hence is not available during the pre-training phase. Although the variational autoencoder can be fine-tuned during the training phase, doing so defeats the purpose of using a BNN.

\subsection{Binary Feature Extraction Phase}
\label{sec:BFEP}
This step is straightforward and takes place on device, the dataset is passed through the trained encoder obtained in the last phase, the output of the encoder can then be stored or directly passed through the binary neural network for training. This step can be done in background on the edge device, when the device is idle as this phase is not that resource intensive.

\subsection{Training Phase}
\label{sec:TP}
The main and final step is training of the binary neural network itself on the edge. The required binary neural network model is trained on the device using the data available. The architecture makes no assumptions about the kind of BNN model or the task required as well as the training methodology. 

Here, the features extracted in the previous step are used to train the binary neural network, the input binarization step is not required as the input space already consists of binary features.

\section{Federated Hybrid BNN}
\label{sec:FedHyBNN}
Federated learning is a distributed privacy-preserving machine learning framework which consists of a single, global model which is trained indirectly using data from numerous asynchronous, heterogeneous devices with the added constraint that data must be stored and processed on-device and no data can be transferred to preserve privacy. Most federated models instead transfer the parameters of locally trained models to the server where these parameters are then aggregated with a federated strategy into a global model. Often, the number of connected edge devices can range in millions. This causes one of the main drawbacks with Federated Learning, the transfer of large number of parameters from the users to the server, whose network bandwidth is usually constrained. 

To solve this problem, we use lightweight BNNs who are really quick to train and also boasts 32x less memory consumption and hence a way to highly reduce the communication costs and scalability problems of federated learning framework as well as make it easier to train a neural network on edge devices.

\begin{figure}[h]
\centering
\includegraphics[width=0.7\textwidth]{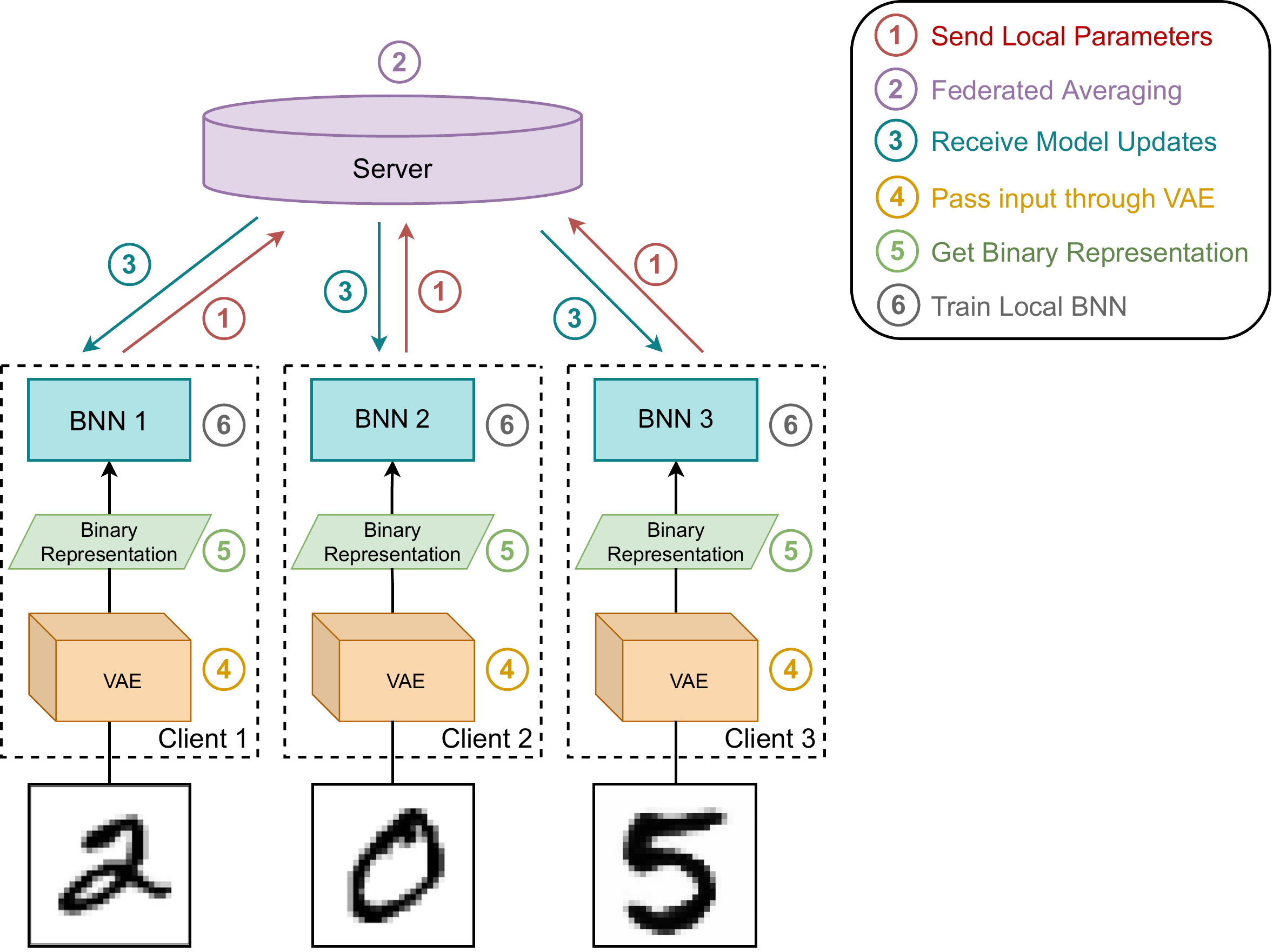}
\caption{Architecture of FedHyBNN}
\label{fig:fedhybnn}
\end{figure}

In this section we discuss the federated counterpart to HyBNN: Federated Hybrid Binary Neural Networks (FedHyBNN). 
The architecture of FedHyBNN is given in Figure \ref{fig:fedhybnn}. Each client consists of the same variational autoencoder but different BNN. During training, only the weights of the BNN are transferred making it much more communication efficient.

The FedHyBNN training can be broken down into six key steps, with three of the steps being identical to HyBNN:

\begin{enumerate}
    \item Representation learning Phase
    \item Local Parameters transfer
    \item Federated Averaging
    \item Model updating
    \item Binary Feature Extraction Phase
    \item Training Phase
\end{enumerate}

The Representation Learning Phase, Binary Feature Extraction Phase and Training Phase are identical to the ones given in Section \ref{sec:RLP}, \ref{sec:BFEP} and \ref{sec:TP} respectively.

During the local parameters transfer phase, the parameters of only the binary neural network of each client are sent to the centralized server for federated averaging and later on updating the model. The received parameters of the BNNs are aggregated together using Federated Averaging (FedAvg) \cite{mcmahan2017communication}, the averaged parameters are then finally transferred back to the clients to update the local models in the Model updating phase.

\section{Experimental Results}
In this section, the proposed HyBNN and FedHyBNN are tested on the MNIST dataset. For the experiment, the VAE of HyBNN follows the same architecture as BinPlay \cite{deja2020binplay}, the encoder consists of three convolutional layers with batch normalization followed by an output layer consisting of 200 neurons. The binary latent space consists of merely 200 bits or 25 bytes. The binary neural network consisted of a one hidden fully connected layer consisting 128 neurons with batch normalization and an output layer with 2 neurons. For binarization, the sign function is used along with the straight-through estimator. 

For the comparison a plain binary convolutional neural network (binary CNN) was chosen which was equivalent to both these models combined, consisting of three convolutional layers followed by a regular fully connected layer with 200 neurons and then immediately followed by one hidden fully connected layer consisting 128 neurons with batch normalization and an output layer with 2 neurons. For the comparison the input was binarized using the same sign function with the straight-through estimator.

The VAE was trained on generating MNIST digits, which comprised of the general task. While there were 5 specific task for the BNN each consisting of distinguishing between two consecutive digits. Task 1 consisted of distinguishing between 0 and 1, Task 2 consisted of distinguishing between 2 and 3 and so on. The VAE in HyBNN was trained for approx. 20 epochs, while the BNN was trained for 100 epochs. The binary CNN was only trained for 15 epochs since the model was overfitting on the data and the test accuracy was decreasing.

\begin{figure}[h]%
    \centering
    \subfloat[\centering Training Accuracy]{{\includegraphics[width=0.45\textwidth]{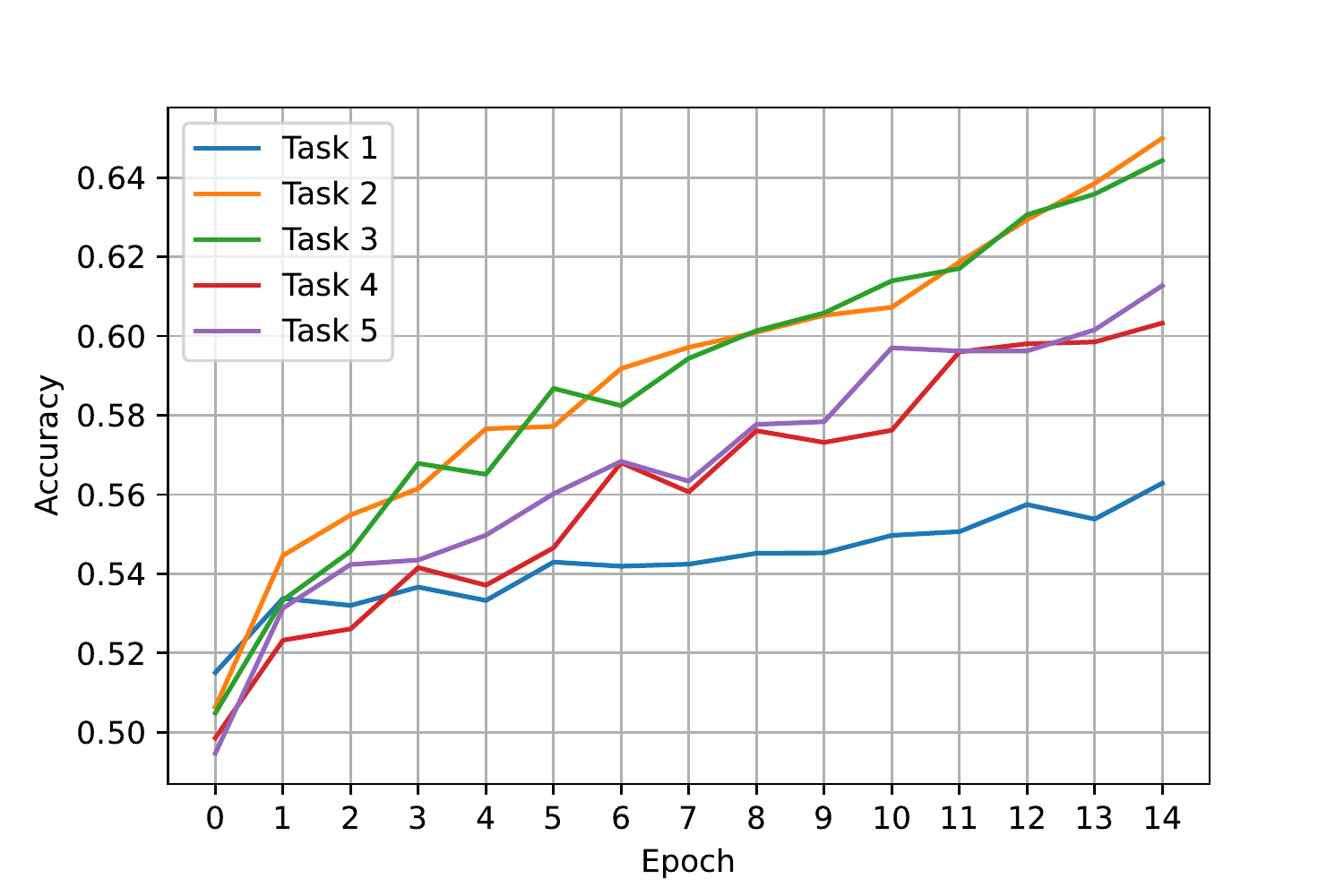} }}%
    \qquad
    \subfloat[\centering Training Loss]{{\includegraphics[width=0.45\textwidth]{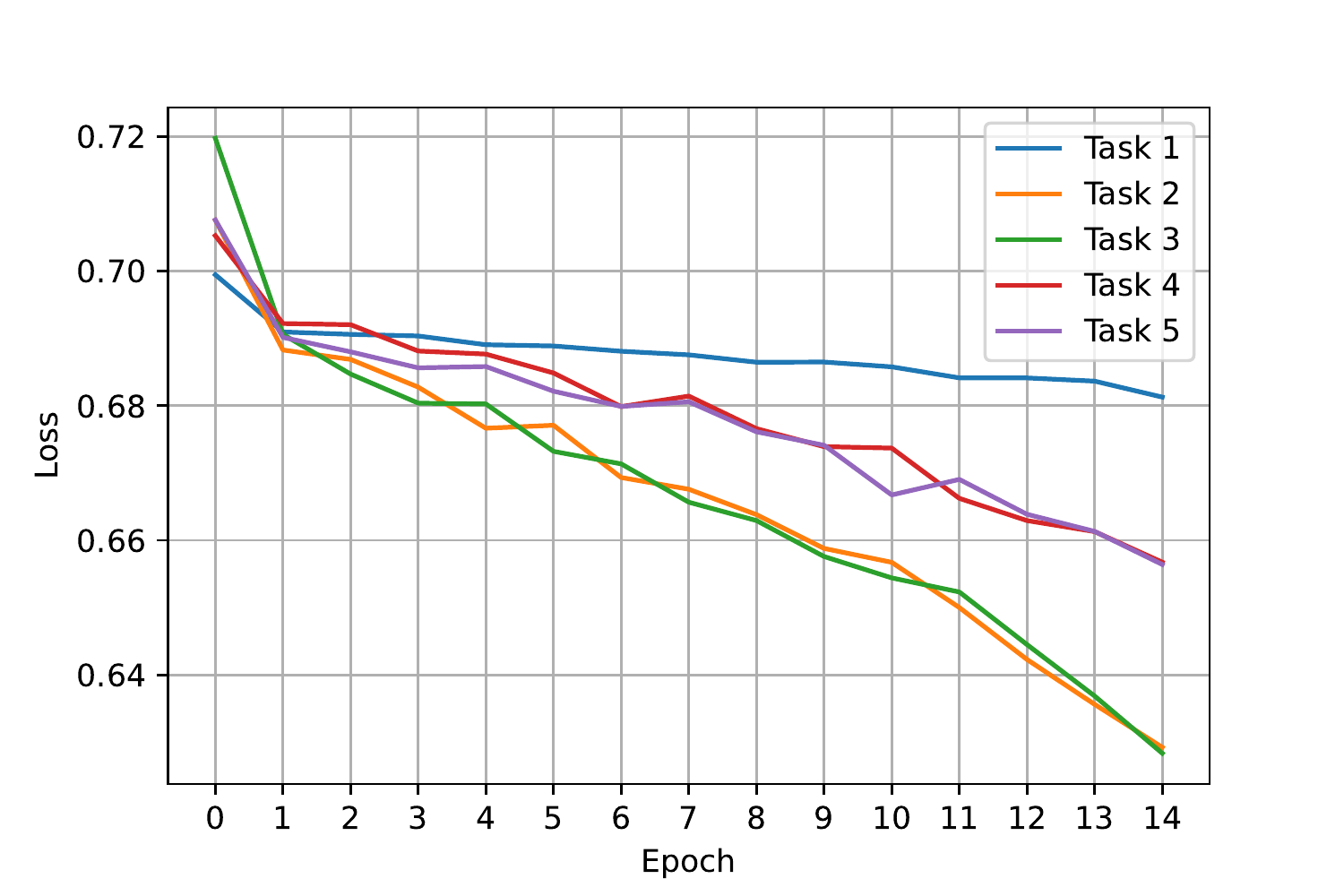} }}%
    \caption{Plain Binary CNN performance}%
    \label{fig:BCNN}%
\end{figure}

The accuracy and loss of the binary CNN is given in Figure \ref{fig:BCNN}, the binary CNN suffered heavily in the tasks, especially task one. The performance of the binary CNN with input binarization was extremely poor, only a little about 50\%, which means the model failed to train, only being as accurate as a random guessing. This is because a majority of information is lost when the input is binarized.

\begin{figure}[h]%
    \centering
    \subfloat[\centering Training Accuracy]{{\includegraphics[width=0.45\textwidth]{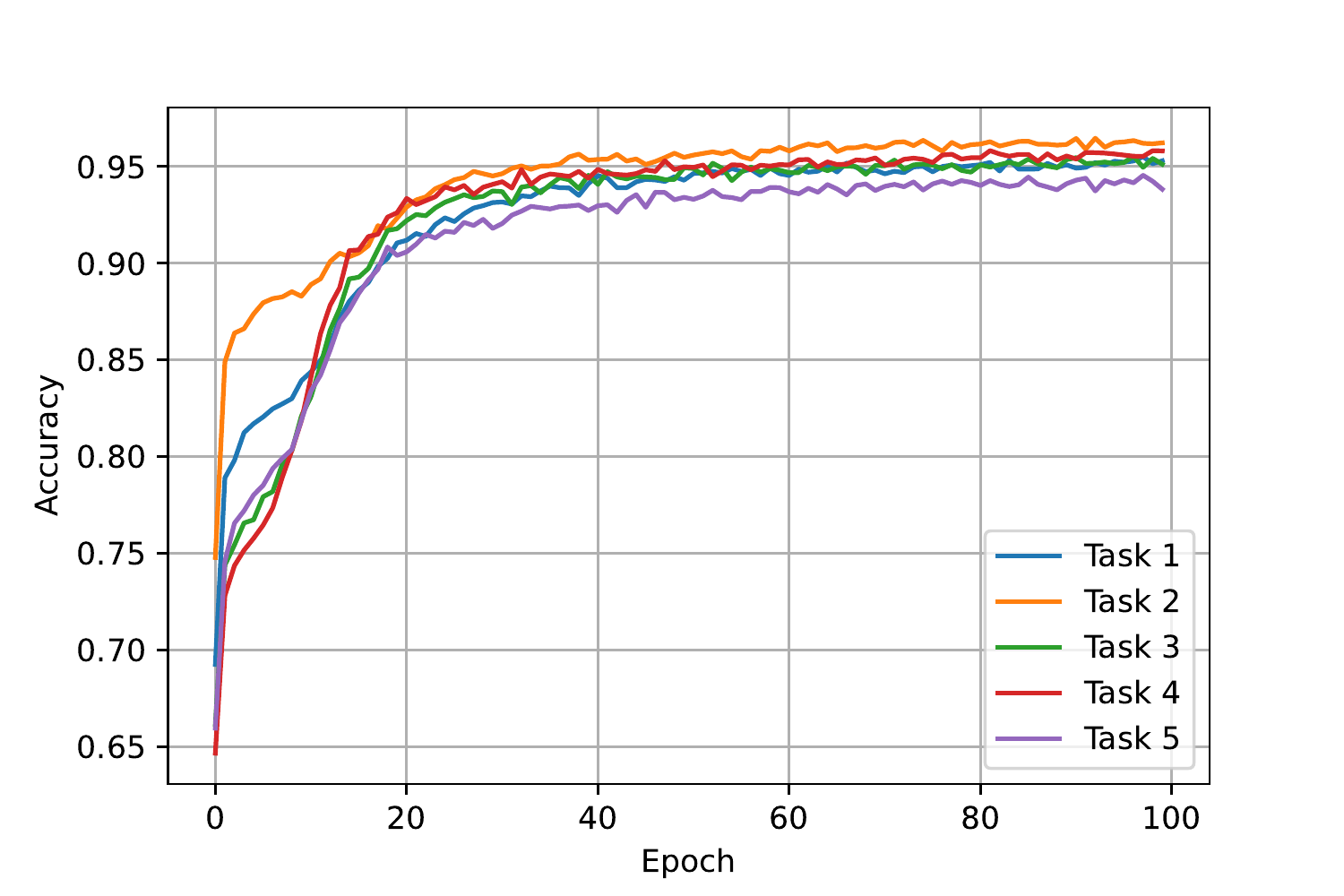} }}%
    \qquad
    \subfloat[\centering Training Loss]{{\includegraphics[width=0.45\textwidth]{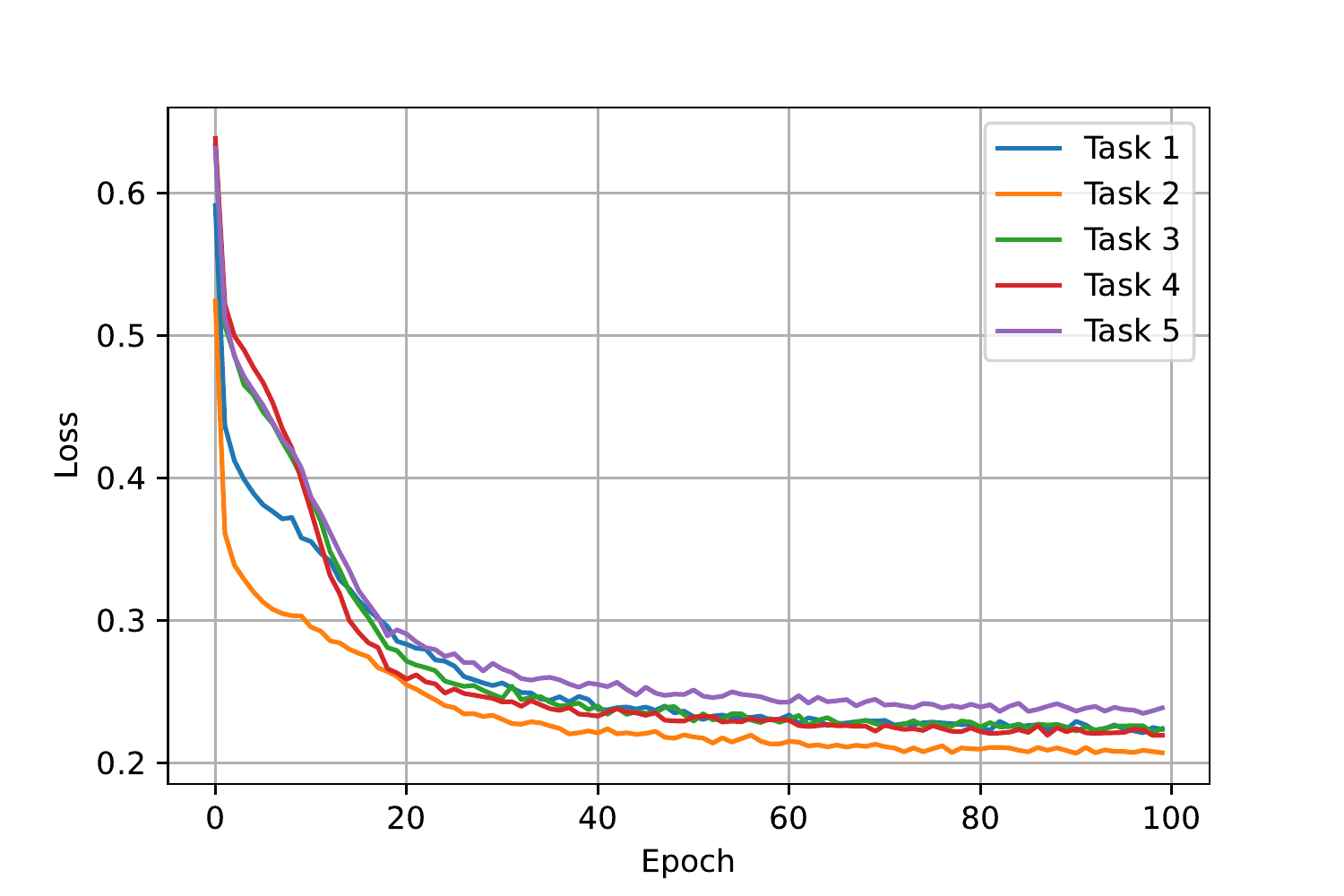} }}%
    \caption{HyBNN Performance}%
    \label{fig:HyBNNper}%
\end{figure}

\begin{figure}[h]%
    \centering
    \subfloat[\centering Test Accuracy]{{\includegraphics[width=0.45\textwidth]{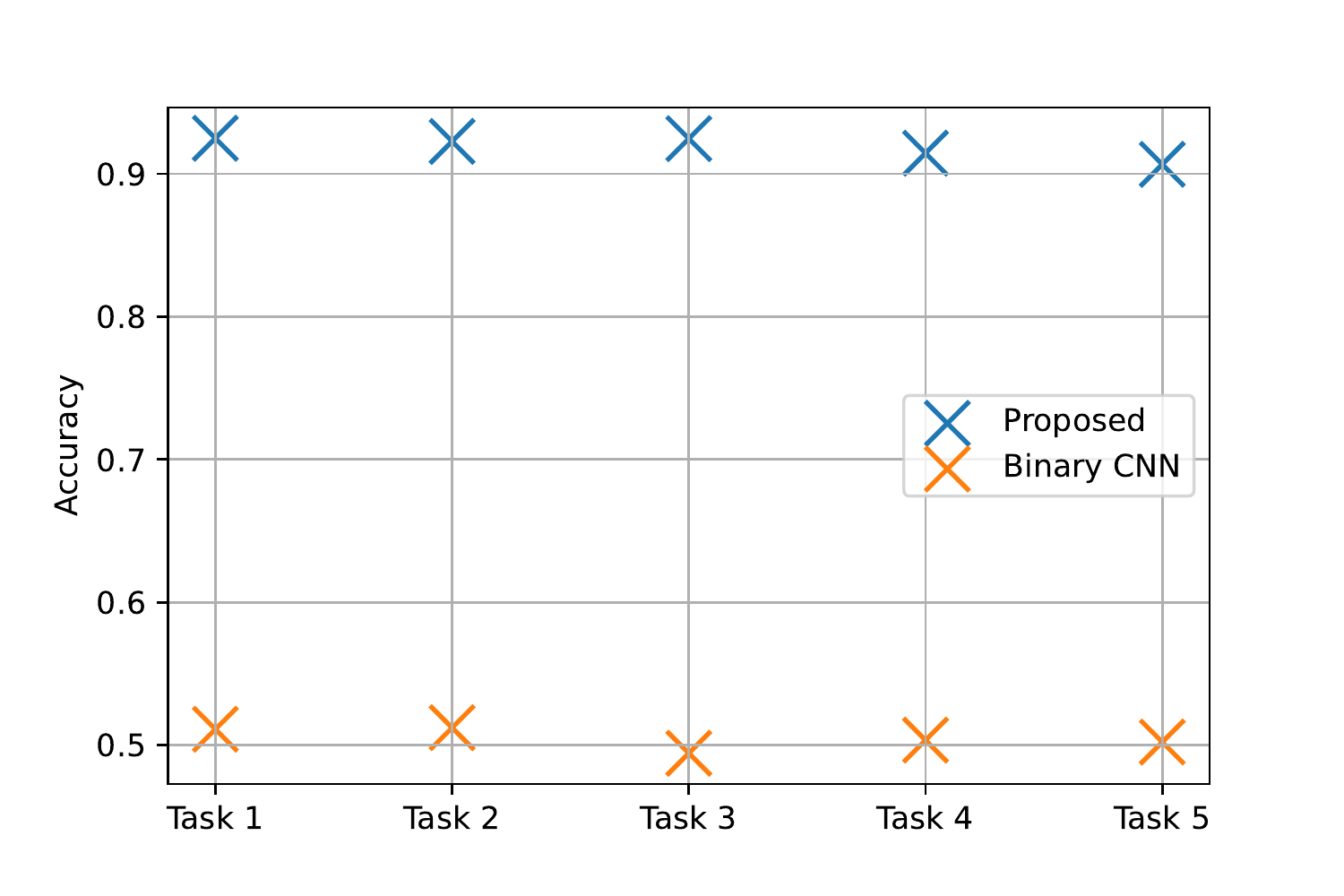} }}%
    \qquad
    \subfloat[\centering Test Loss]{{\includegraphics[width=0.45\textwidth]{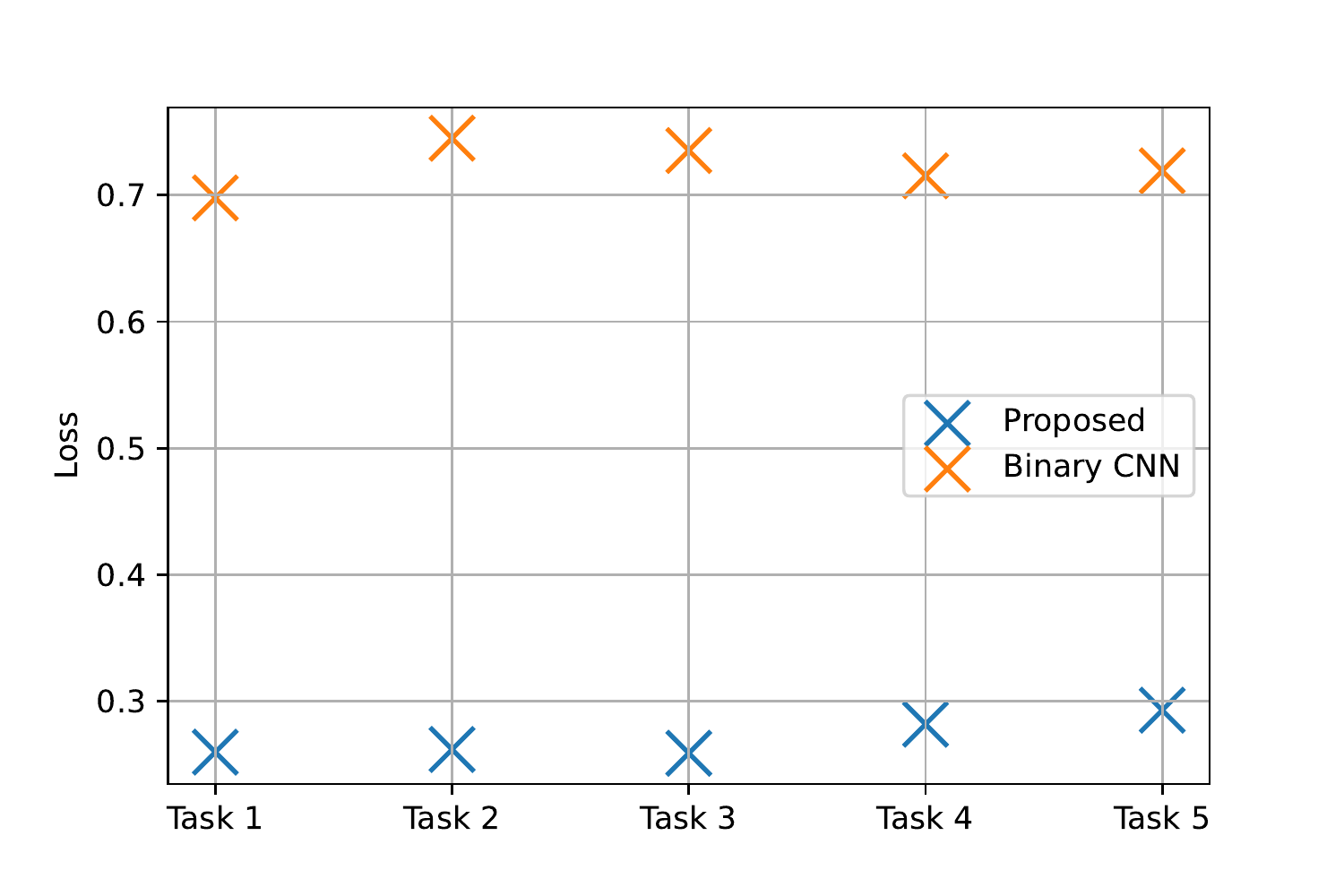} }}%
    \caption{Comparison}%
    \label{fig:comparison}%
\end{figure}

On the other hand our proposed HyBNN model was able to boost up the accuracy to over 95\%, the accuracy and loss for all five tasks is given in Figure \ref{fig:HyBNNper}. The comparison between accuracies for each task is given in Figure \ref{fig:comparison}. As described there is a considerable increase in performance (more than 40\%).

\begin{figure}[h]%
    \centering
    \subfloat[\centering Training Accuracy]{{\includegraphics[width=0.45\textwidth]{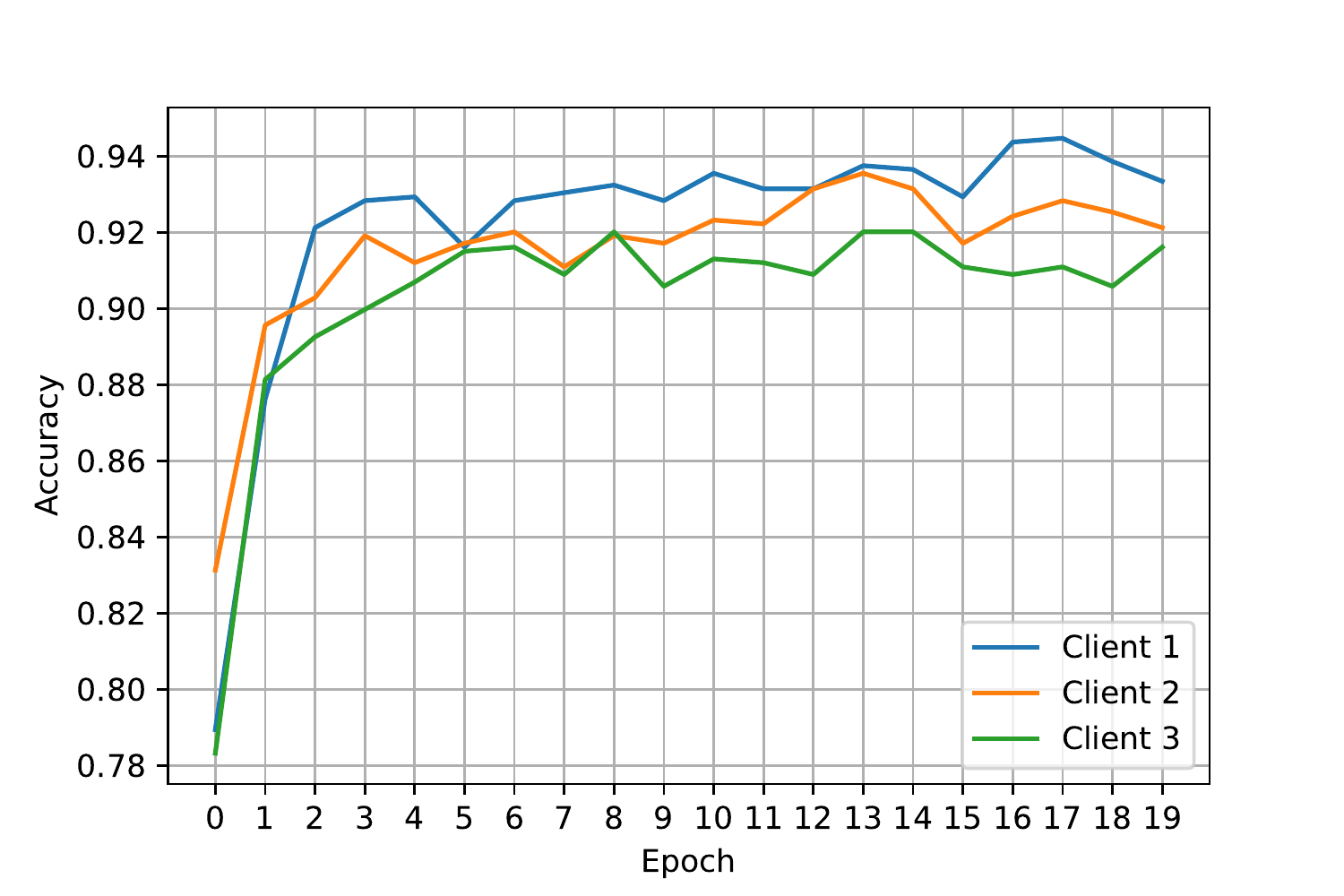} }}%
    \qquad
    \subfloat[\centering Training Loss]{{\includegraphics[width=0.45\textwidth]{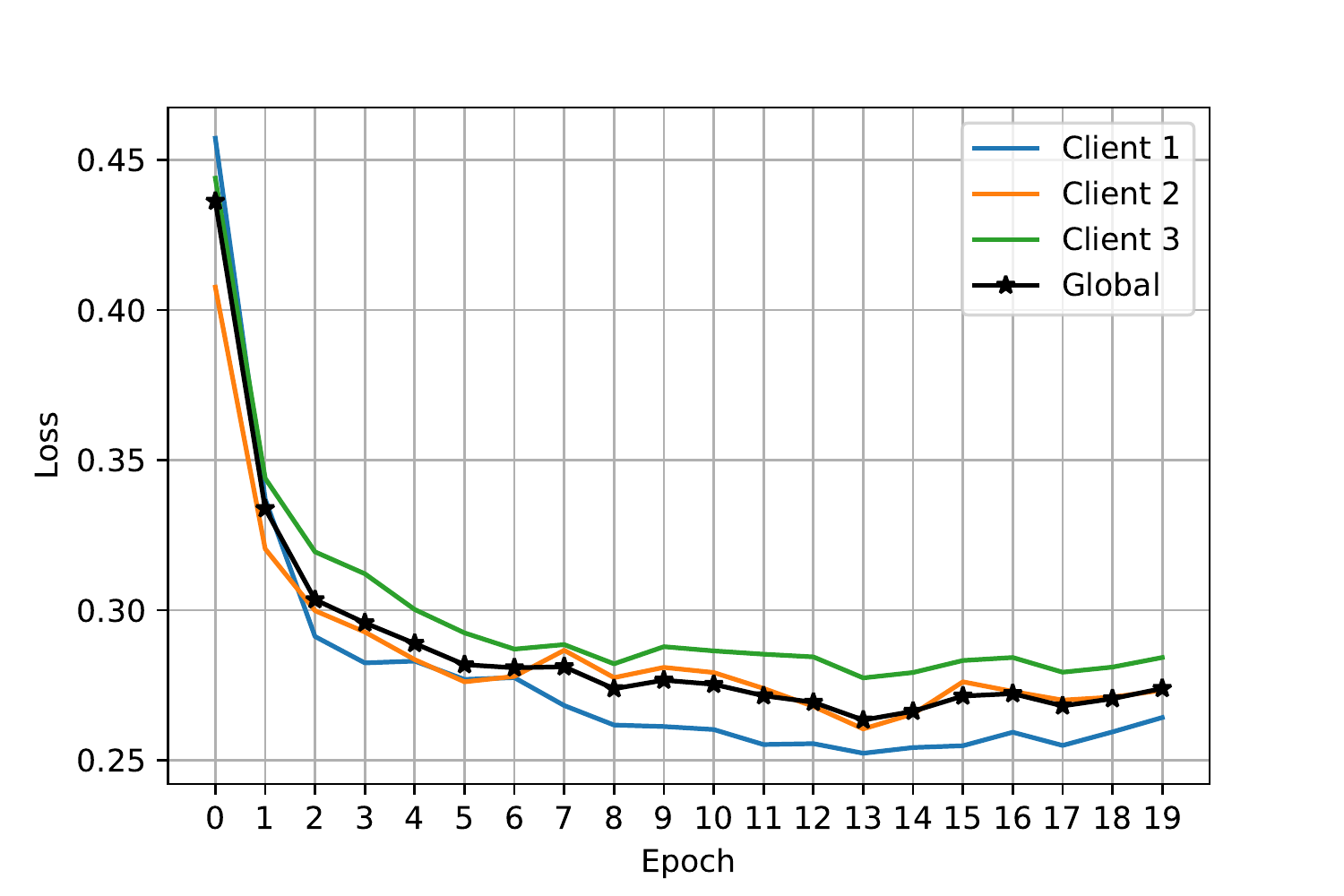} }}%
    \caption{Federated Performance}%
    \label{fig:fl_results}%
\end{figure}

For FedHyBNN, we implemented a three client scenario to demonstrate the performance of our proposed model. The clients were able to converge quickly and in under 20 epochs they were able to reach the same accuracy as the centralized HyBNN. The results are given in Figure \ref{fig:fl_results}.

\vskip\baselineskip
\vskip\baselineskip
\vskip\baselineskip

\section{Limitations}

Although the architecture shows considerable potential, it still has some limitations. The performance of the proposed model decreased (although was still greater than the plain binary CNN) if it was trained on all five tasks consecutively, like in a continual learning setup. More research needs to be done in the use of continual learning with HyBNN. The binary representation seemed to work best for binary classification. The performance can be probably be improved as more work is done into designing other models to generate the binary representation.

\section{Conclusion and Future Work}
In this paper, we introduced a novel hybrid full-precision and binary neural network architecture. To our knowledge, this is the first example of a hybrid neural network that combines the advantages of binary neural networks with the high accuracy of full-precision deep neural networks. We established that it is possible to leverage the VAE as a binary feature extractor to increase the accuracy of BNNs close to its full-precision counterpart. We have shown that the HyBNN architecture can improve the accuracy of existing BNNs by a huge margin. We also introduced FedHyBNN, the federated counterpart to HyBNN that offers extremely communication efficient federated learning in real time. Since, our architecture makes no assumptions about the type of BNN, the algorithm to train it or the given task, it can be applied to a wide variety of use cases.  Future work include application of the architecture to other tasks and experimentation with continual learning.

\bibliography{refs}

\end{document}